\pdfoutput=1

\documentclass[11pt]{article}

\usepackage[final]{coling}

\usepackage{times}
\usepackage{latexsym}

\usepackage[T1]{fontenc}

\usepackage[utf8]{inputenc}

\usepackage{microtype}

\usepackage{inconsolata}

\usepackage{graphicx}
\usepackage{booktabs}
\usepackage{caption}
\usepackage{wrapfig}
\usepackage{multirow}
\usepackage{makecell}
\usepackage{xcolor}
\usepackage{color}
\usepackage{tcolorbox}
\usepackage{colortbl}
\usepackage{amsmath}
\usepackage{enumitem}
\newcommand{\ignore}[1]{}
\newcommand{\ie}{\emph{i.e.,} }

\newcommand{\eg}{\emph{e.g.,} }
\newcommand{\etc}{\emph{etc}}

\title{What Makes for Good Visual Instructions? Synthesizing Complex Visual Reasoning Instructions for Visual Instruction Tuning} 

\author{
Yifan Du$^{1}$\thanks{Equal Contribution.},
Hangyu Guo$^{1*}$,
Kun Zhou$^{2*}$,
Wayne Xin Zhao$^{1}$\thanks{Corresponding Author},
Jinpeng Wang$^3$,\\
\textbf{Chuyuan Wang$^3$},
\textbf{Mingchen Cai$^3$},
\textbf{Ruihua Song$^1$},
\textbf{Ji-Rong Wen$^{1}$}\\
[2mm]
$^1$~Gaoling School of Artificial Intelligence, Renmin University of China\\
$^2$~School of Information, Renmin University of China \quad
$^3$~Meituan Group\\
{\tt \{yifandu1999, hyguo0220, batmanfly\}@gmail.com},
{\tt francis\_kun\_zhou@163.com}
}

\begin{document}
\maketitle
\begin{abstract}
Visual instruction tuning is crucial for enhancing the zero-shot generalization capability of Multi-modal Large Language Models (MLLMs). In this paper, we aim to investigate a fundamental question: ``\textit{what makes for good visual instructions}''. Through a comprehensive empirical study, we find that instructions focusing on complex visual reasoning tasks are particularly effective in improving the performance of MLLMs, with results correlating to instruction complexity. Based on this insight, we develop a systematic approach to automatically create high-quality complex visual reasoning instructions. Our approach employs a \textit{synthesize-complicate-reformulate} paradigm, leveraging multiple stages to gradually increase the complexity of the instructions while guaranteeing quality. Based on this approach, we create the \textbf{ComVint} dataset with 32K examples, and fine-tune four MLLMs on it. Experimental results consistently demonstrate the enhanced performance of all compared MLLMs, such as a 27.86\% and 27.60\% improvement for LLaVA on MME-Perception and MME-Cognition, respectively. Our code and data are publicly available at the link: \url{https://github.com/RUCAIBox/ComVint}.

\end{abstract}

\section{Introduction}
\label{sec:intro}


To extend the application scope of Large Language Models (LLMs)~\cite{Zhao-2023-arxiv-survey, brown-2020-nips-language} , a surge of work~\cite{liu-2023-arxiv-visual, ye-2023-arxiv-mplug} augments LLMs with vision encoders to endow the ability of multi-modal cognition and reasoning, leading to the emergence of Multi-modal Large Language Models~(MLLMs)~\cite{yin-2023-arxiv-survey, li-2023-arxiv-multimodal}. To achieve good performance, most work first pre-trains the MLLM on a large collection of image-text pairs~(\eg LAION~\cite{schuhmann-2021-arxiv-LAION} and CC~\cite{Changpinyo-2021-cvpr-conceptual}) to align the text and visual representations, and then fine-tunes it on visual instructions to improve the zero-shot generalization capability~\cite{liu-2023-arxiv-visual, zhang-2023-arxiv-xcomposer}.

\begin{figure}
  \centering
    \includegraphics[width=0.98\linewidth]{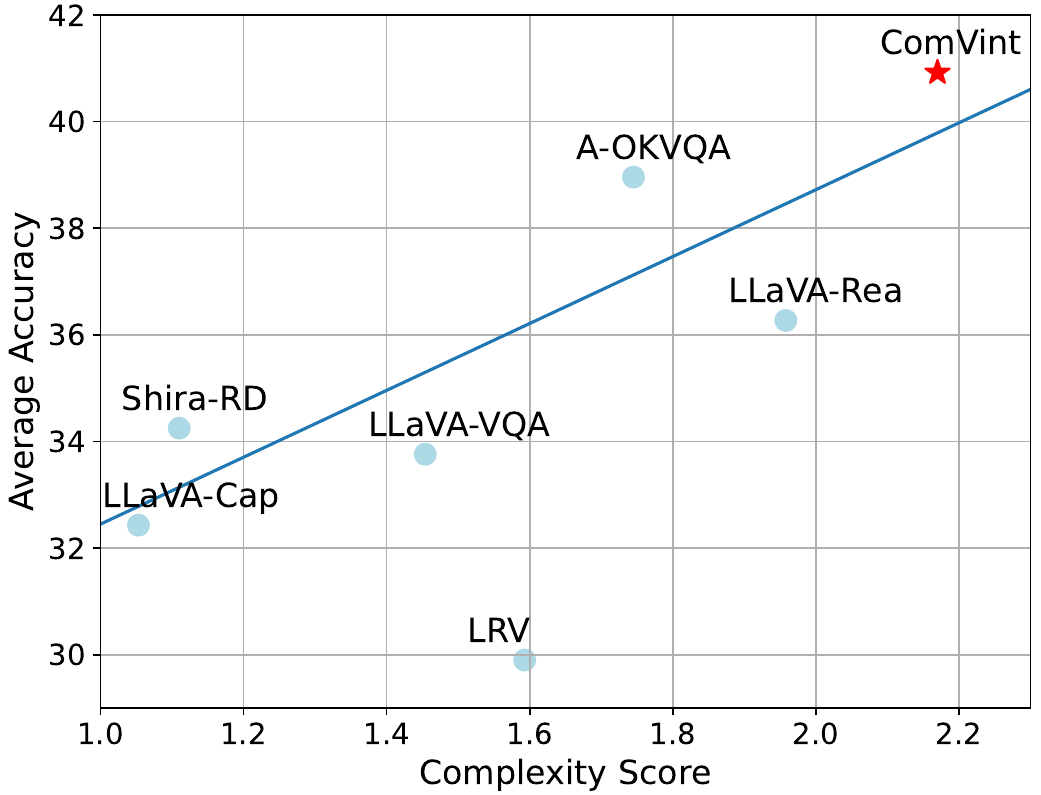}
    \caption{The relation between the complexity of the used instruction set and the average performance of four models on SEED-Bench and MME. The complexity is measured by the reasoning steps counted by ChatGPT.}
    \label{fig:complexity_score}
\end{figure}

A visual instruction typically consists of an image, a task description, and a text output~\cite{liu-2023-arxiv-visual, yin-2023-arxiv-survey}. 
Great efforts have been made to construct high-quality visual instruction datasets, including collecting existing datasets~\cite{Li-2023-arxiv-m3it, li-2023-arxiv-mimic} or synthesis via LLMs~\cite{liu-2023-arxiv-aligning, chen-2023-arxiv-shikra}. Despite the prosperity, there is still a lack of a systematic comparison of instruction sets in terms of effectiveness, \emph{based on the same settings} of the backbone model and training strategies. Thus, it remains unclear which instruction sets are more effective and what factors contribute to good instruction data.

Considering this issue, we would like to investigate a more fundamental question, \ie  ``\emph{what makes for good visual instructions}''. For this purpose, we first conduct a comprehensive evaluation of existing visual instruction sets, aiming to identify the key factors that contribute to effective instructions for MLLMs. Specifically, based on six representative instruction datasets and two popular MLLM, we mainly consider examining two important aspects, namely task types and instruction characteristics. 
According to the empirical study, we have two main findings:

\begin{itemize}[left=0pt]
    \item The visual reasoning task is more helpful in boosting the performance than image captioning and visual question answering tasks.

    \item Increasing the instruction complexity is more helpful to improve the performance, than enhancing instruction diversity and integrating fine-grained spatial information.
\end{itemize}

Additionally, as shown in Figure~\ref{fig:complexity_score}, as the complexity of visual instruction datasets increases, the average benchmark performance is also consistently improved, following an approximate linear trend (details in Section~\ref{sec:experiments}).
All the above results motivate us to construct complex visual reasoning instructions to enhance MLLMs. 
However, it is hard to directly prompt GPT-4~\cite{open-2023-arxiv-gpt4} for synthesizing sufficiently complex and non-hallucination visual instructions~\cite{li-2023-arxiv-evaluating}.
To address this, we developed a systematic multi-stage pipeline to gradually enhance the quality and complexity of the generated instructions. 
Concretely, our approach adopts a \emph{synthesize-complicate-reformulate} pipeline to generate the instruction, where corresponding prompts are devised to guide GPT-4.
In the complication stage, we guide GPT-4 to fully utilize both image content and outside knowledge\footnote{According to~\cite{marino-2019-cvpr-okvqa}, outside knowledge refers to the knowledge that is not provided by the image, \eg inferring the latitude of a location in the image.} to improve the complexity, and iteratively verify the accuracy of the instruction to ensure the quality. 
Finally, we reformulate it into multiple formats for better adaptation to various downstream tasks.

\ignore{In contrast to existing work, our approach mainly focuses on automatically synthesizing high-quality complex visual reasoning instruction, to enrich the data resource in the research community for training more effective MLLMs.}



Using the above approach, we synthesize a \textbf{Com}plex \textbf{V}isual reasoning \textbf{in}s\textbf{t}ruction dataset, namely \textbf{ComVint}, consisting of 32K examples, and fine-tune four representative MLLMs~(\ie BLIP-2, LLaVA, MiniGPT-4, and InstructBLIP) on it.
Evaluation results on two comprehensive benchmarks, SEED-Bench~\cite{li-2023-arxiv-SEED} and MME~\cite{fu-2023-arxiv-MME}, demonstrate that our instruction dataset significantly enhances the performance of these MLLMs, outperforming existing visual instruction collections.
For instance, leveraging our dataset leads to a remarkable improvement of 27.86\% and 27.60\% in the performance of LLaVA on MME-Perception and MME-Cognition, respectively. 
\section{Background}

\label{sec:preliminary}
\paragraph{Multi-modal Large Language Models.}
Multi-modal Large Language Models~(MLLMs)~\cite{li-2023-arxiv-multimodal} are advanced generative models capable of processing information from various modalities~(\eg image, video, and audio) and generating corresponding textual responses. This work focuses on MLLMs in the visual modality, typically consisting of an image encoder, an LLM, and a connection module. The image is first encoded into patch embeddings by the image encoder and the connection module, then concatenated with text embeddings, enabling the LLM to comprehend the image and generate the response auto-regressively. MLLMs undergo vision-language pre-training to align the vision encoder and LLM, followed by visual instruction tuning to enhance instruction following and understanding ability~\cite{liu-2023-arxiv-visual, Zhang-arxiv-2023-internlm}.

\paragraph{Visual Instruction Tuning.}
Instruction tuning~\cite{wei-2022-iclr-finetuned, chung-2022-arxiv-scaling} is important to improve the ability of LLMs in instruction following and generalization on unseen tasks~\cite{longpre-2023-arxiv-the, wang-2023-arxiv-how}. It employs a text-formatted task description and the expected outcome to fine-tune LLMs in a supervised way. Inspired by the success of LLMs, instruction tuning has been adapted to develop MLLMs for visual tasks, termed \emph{visual instruction tuning}~\cite{zhu-2023-arxiv-MiniGPT,liu-2023-arxiv-visual}. Typically, a visual instruction comprises an image $X_{I}$, a textual task instruction $X_{T}$, and a corresponding output text $Y_{T}$. During training, MLLMs learn to generate $Y_{T}$ conditioned on $X_{I}$ and $X_{T}$. 


\ignore{To improve the performance of a given MLLM, there are two typical approaches: improving the quality of visual instructions or improving the training method~\cite{liu-2023-arxiv-improved,chen-2023-arxiv-minigptv2,zhai-2023-arxiv-investigating,sun-2023-arxiv-aligning}. In this work, we focus on exploring which kinds of visual instructions are more useful to improve MLLMs, and leave the improved training method as future work.}

\subsection{Visual Instruction Collections}
\label{sec:visual_instruction_collections}


To get a sense of what constitutes good visual instructions, we review and categorize existing collections in Table~\ref{tab:intro}. Broadly, visual instructions are crafted to address specific tasks and incorporate various considerations (\eg diversity and complexity). Thus, we discuss prior efforts in two major aspects: task types and instruction characteristics.

\paragraph{Task Types.} Most visual instruction datasets~\cite{liu-2023-arxiv-visual, Li-2023-arxiv-m3it} are derived from existing multi-modal datasets and primarily focus on three types of tasks:

\begin{itemize}[left=0pt]
    \item \textit{Image captioning}: it requires the model to generate a free-form description of an image. 

    \item \textit{Visual question answering~(VQA)}: it requires the model to answer a question about the image, \eg counting the objects and recognizing the color. 

    \item \textit{Visual reasoning}: it requires the model to perform reasoning based on the image context, \eg conjecturing the relationship between two objects, and answering questions involving commonsense reasoning. 
\end{itemize}

We examine the effects of task type using the LLaVA-Instruct~\cite{liu-2023-arxiv-visual}, which includes 23K image captions, 58K conversations, and 77K visual reasoning questions. We divide it into three subsets corresponding to the three task types, namely {LLaVA-Caption}, {LLaVA-VQA}, and {LLaVA-Reasoning}, respectively.

\paragraph{Instruction Characteristics.}
In addition to the task types, recent studies~\cite{liu-2023-arxiv-aligning, chen-2023-arxiv-shikra, zhang-2023-arxiv-gpt4roi, chen-2023-arxiv-pvit} also attempt to endow visual instruction collection with special characteristics to further improve the performance of MLLMs. 

\begin{itemize}[left=0pt]
    \item \textit{Task diversity:} existing work~\cite{wei-2022-iclr-finetuned,liu-2023-arxiv-aligning} has found that increasing the task diversity can improve the zero-shot ability for task solving. This can typically be achieved by aggregating instructions from different tasks.
    
    \item \textit{Instruction complexity:} enhancing instruction complexity is a widely used strategy to improve the performance of LLMs~\cite{xu-2023-arxiv-wizardlm}, and we can also utilize complex multi-modal tasks (\eg multi-hop cross-modal reasoning) to improve the performance of MLLMs.

    \item \textit{Fine-grained spatial information:} it is important for MLLMs to recognize fine-grained spatial details of objects in an image. For this purpose, the spatial coordinates annotations can be included in the textual instructions~\cite{chen-2023-arxiv-shikra,chen-2023-arxiv-pvit}.
    
\end{itemize}

To study the effect of these characteristics, we select {LRV}~\cite{liu-2023-arxiv-aligning}, {A-OKVQA}~\cite{schwenk-2022-eccv-aokvqa}, and {Shikra-RD}~\cite{chen-2023-arxiv-shikra}, three representative instruction sets with diverse task types, complex outside knowledge, and fine-grained spatial information, respectively.

\begin{table}[t]
\small
    \centering
    
    \setlength{\tabcolsep}{10pt} 
    \scalebox{0.8}{\begin{tabular}{llll}
    \toprule
         \makecell[l]{Instruction}&  Number&  Task Type& Characteristics\\
         \midrule
         \makecell[l]{LLaVA-Cap}&  23K&  Cap&$\backslash$ \\ 
         LLaVA-VQA$^*$&  256K&  VQA&$\backslash$ \\ 
         \makecell[l]{LLaVA-Rea}&  77K&  Rea& $\backslash$\\ 
         LRV&  150K&  \makecell[l]{Cap, VQA, Rea}& Diverse\\
         Shikra-RD&  4K&  \makecell[l]{Rea}& \makecell[l]{Fine-grained}\\
         ComVint (Ours)&  32K&  Rea& Complex\\
    \bottomrule
    \end{tabular}}
    \caption{Comparison of existing synthesized visual instruction collections. ``Cap'' shorts for Caption and ``Rea'' shorts for Reasoning. *We divide the multi-turn conversation in LLaVA into individual questions, resulting in 256K VQA instructions.}
    \label{tab:intro}
\end{table}
\section{Empirical Analysis of Visual Instructions}
\begin{table*}[t]
    \centering
    \scalebox{0.8}
    {\begin{tabular}{cl|cccc|cccc}
    \toprule
 \multicolumn{2}{c|}{Baseline Model} &\multicolumn{4}{c|}{MiniGPT-4}&\multicolumn{4}{c}{BLIP-2}\\
    \midrule
           \multicolumn{2}{c|}{Benchmark}&  \makecell[c]{SEED-Bench \\Image ACC}&  \makecell[c]{MME-P\\ ACC+}&  \makecell[c]{MME-C\\ ACC+}&  Average& \makecell[c]{SEED-Bench \\Image ACC}& \makecell[c]{MME-P\\ ACC+}& \makecell[c]{MME-C\\ ACC+}&Average\\
    \midrule
           &Original&  43.31&  26.96&  10.77&  27.01&  53.68& 41.25 & 15.38 &36.77 \\
    \midrule
   \multirow{3}{*}{\makecell[c]{A}}&+LLaVA-Cap&  \cellcolor[HTML]{D9E9F3}41.60&  \cellcolor[HTML]{92BFDB}10.12&  \cellcolor[HTML]{D9E9F3}1.54 &  \cellcolor[HTML]{C4DDEC}17.75& \cellcolor[HTML]{D9E9F3}52.31& \cellcolor[HTML]{92BFDB}28.38 & \cellcolor[HTML]{C4DDEC}11.54 &\cellcolor[HTML]{92BFDB}30.74 \\
           &+LLaVA-VQA&  \cellcolor[HTML]{FEE0D2}45.97&  \cellcolor[HTML]{92BFDB}12.87 &  \cellcolor[HTML]{92BFDB}5.38 &  \cellcolor[HTML]{92BFDB}21.41& \cellcolor[HTML]{C4DDEC}51.00& \cellcolor[HTML]{92BFDB}31.88 & \cellcolor[HTML]{D9E9F3}14.62 &\cellcolor[HTML]{C4DDEC}32.50 
\\
           &+LLaVA-Rea&  \cellcolor[HTML]{FEE0D2}43.69&  \cellcolor[HTML]{FC8D59}\textbf{40.59} &  \cellcolor[HTML]{FC8D59}16.15 &  \cellcolor[HTML]{FC8D59}33.48& \cellcolor[HTML]{C4DDEC}50.94& \cellcolor[HTML]{FEE0D2}43.33 & \cellcolor[HTML]{FC8D59}20.77 &\cellcolor[HTML]{FEE0D2}38.35 
\\
\midrule
           \multirow{3}{*}{\makecell[c]{B}}&+LRV&  \cellcolor[HTML]{FC8D59}\textbf{50.92}&  \cellcolor[HTML]{92BFDB}3.12&  \cellcolor[HTML]{92BFDB}0.77 &  \cellcolor[HTML]{92BFDB}18.27& \cellcolor[HTML]{FEE0D2}\textbf{54.64}& \cellcolor[HTML]{92BFDB}10.88 & \cellcolor[HTML]{92BFDB}5.38 &\cellcolor[HTML]{92BFDB}23.63 \\
           &+Shikra-RD&  \cellcolor[HTML]{C4DDEC}41.75&  \cellcolor[HTML]{92BFDB}8.33 &  \cellcolor[HTML]{92BFDB}1.54 &  \cellcolor[HTML]{92BFDB}17.21& \cellcolor[HTML]{D9E9F3}52.92& \cellcolor[HTML]{C4DDEC}36.05 & \cellcolor[HTML]{FEE0D2}16.15 &\cellcolor[HTML]{D9E9F3}35.04 \\
            &+A-OKVQA&  \cellcolor[HTML]{FEE0D2}43.99&  \cellcolor[HTML]{FCA982}36.71&  \cellcolor[HTML]{FC8D59}\textbf{21.54}&  \cellcolor[HTML]{FC8D59}\textbf{34.08}& \cellcolor[HTML]{D9E9F3}52.60& \cellcolor[HTML]{FCA982}\textbf{46.83} & \cellcolor[HTML]{FC8D59}\textbf{24.62} &\cellcolor[HTML]{FC8D59}\textbf{41.35} \\
 
\bottomrule
    \end{tabular}}
    \caption{The results on SEED-Bench and MME after fine-tuning MiniGPT-4 and BLIP-2 using different instruction collections. MME-P and MME-C short for MME-Perception and MME-Cognition, respectively. Cells shaded in orange indicate that fine-tuning enhances the performance, while blue indicates performance degradation.}
    \label{tab:empirical}
\end{table*}
\label{sec:empirical}
In this section, we empirically study the effect of different task types and different in visual instruction tuning by fine-tuning two representative MLLMs~(\ie BLIP-2~\cite{Li-2023-ICML-BLIP2} and MiniGPT-4~\cite{zhu-2023-arxiv-MiniGPT}) on the  visual instruction collections selected in Section~\ref{sec:preliminary}.

\subsection{Experiment Setup}
\label{sec:empirical_setup}

\paragraph{Backbone MLLMs.} We select two models with minimal or no instruction tuning to clearly study the effect of different factors: 
\ignore{Existing work mostly performs visual instruction tuning on the backbone MLLMs that have been either {pre-trained on large-scale image-text pairs} or {fine-tuned on basic instructions} (\eg image captioning). For the two settings, we choose BLIP-2~\cite{Li-2023-ICML-BLIP2} and MiniGPT-4~\cite{zhu-2023-arxiv-MiniGPT} as the representative backbone models, respectively.}

\begin{itemize}[left=0pt]
    \item \emph{BLIP-2}: it incorporates a lightweight querying Transformer to connect a fixed vision encoder and a fixed LLM. It is only pre-trained on large-scale image-text pairs, and has not been fine-tuned with visual instructions.
    \item \emph{MiniGPT-4}: it employs the similar architecture as BLIP-2 and adopts Vicuna as the LLM. It is first pre-trained on 5M image-text pairs and then fine-tuned on 3,500 image captions.
\end{itemize}

We follow the training strategies in~\cite{zhu-2023-arxiv-MiniGPT, Li-2023-ICML-BLIP2}, fine-tuning the Q-Former in BLIP-2 and the linear layer between the Q-Former and LLM in MiniGPT-4. \ignore{\textcolor{red}{Although much work has conducted similar empirical evaluation, they mainly compare the performance of different MLLMs trained on specific instruction collections~\cite{xu-2023-arxiv-lvlm, li-2023-arxiv-reform}. In contrast, we would like to compare the effect of instruction collections on the same backbone configuration and training procedure.}}

\paragraph{Evaluation Benchmark.}
We select two widely-used benchmarks, \ie MME~\cite{fu-2023-arxiv-MME} and SEED-Bench~\cite{li-2023-arxiv-SEED} for evaluation:
\begin{itemize}[left=0pt]
    \item \emph{MME}: it aims to measure the perception and cognition abilities of MLLMs. Each instance comprises one image and two questions. Following~\cite{fu-2023-arxiv-MME}, we report the accuracy of whether both questions of an instance are answered correctly, denoted as \emph{ACC+}. 
    
    \item \emph{SEED-Bench}: it contains 12 tasks to evaluate the image and video understanding capacity of MLLMs. Since most MLLMs do not consider video understanding ability, we only evaluate the models on image understanding tasks. Following~\cite{zhang-2023-arxiv-xcomposer}, we employ accuracy in image understanding tasks as the evaluation metric and denote it as \emph{Image ACC}. 
\end{itemize}

\subsection{Results and Analysis}
\label{sec:empirical_results_analysis}
We categorize the results into two groups based on the task types and special characteristics of the instructions: (A) includes LLaVA-Cap (image captioning), LLaVA-VQA (visual question answering), and LLaVA-Rea (visual reasoning); (B) includes LRV (diversity), A-OKVQA (complexity), and Shikra-RD (spatial annotation). We can obtain the following insights from the results in Table~\ref{tab:empirical}.

\emph{Finding 1: among task types in group A, the visual reasoning task yields the best performance compared to image captioning and visual question answering.} Concretely, fine-tuning MLLMs on the LLaVA-Cap leads to a noticeable performance degradation across all benchmarks, while LLaVA-Rea results in significant improvements, with LLaVA-VQA results showing intermediate results in most cases. This suggests that the performance advantage may correlate positively with the difficulty of visual instructions.

\emph{Finding 2: for instruction characteristics in group B, complexity is more important than task diversity and fine-grained information.} Concretely, MLLMs fine-tuned on A-OKVQA achieve the highest accuracy across all benchmarks, emphasizing the pivotal role of complex instructions in boosting model performance. In contrast, LRV and Shikra-RD show minimal or negative impacts, indicating limited benefits from increased task diversity or spatial details.

To summarize, reasoning-oriented~(LLaVA-Rea) and complexity-enhanced~(A-OKVQA) instruction sets are particularly useful in improving the performance of MLLMs in our experiments. However, LLaVA-Rea still has limited complexity. Though A-OKVQA contains complex task instructions, it is mainly constructed by human annotators, and the instruction complexity is limited to the annotator's abilities. Therefore, it is desirable to develop automatic approaches to produce complex visual reasoning instructions at scale.
\begin{figure*}[t]
    \centering
    \includegraphics[width=\textwidth]{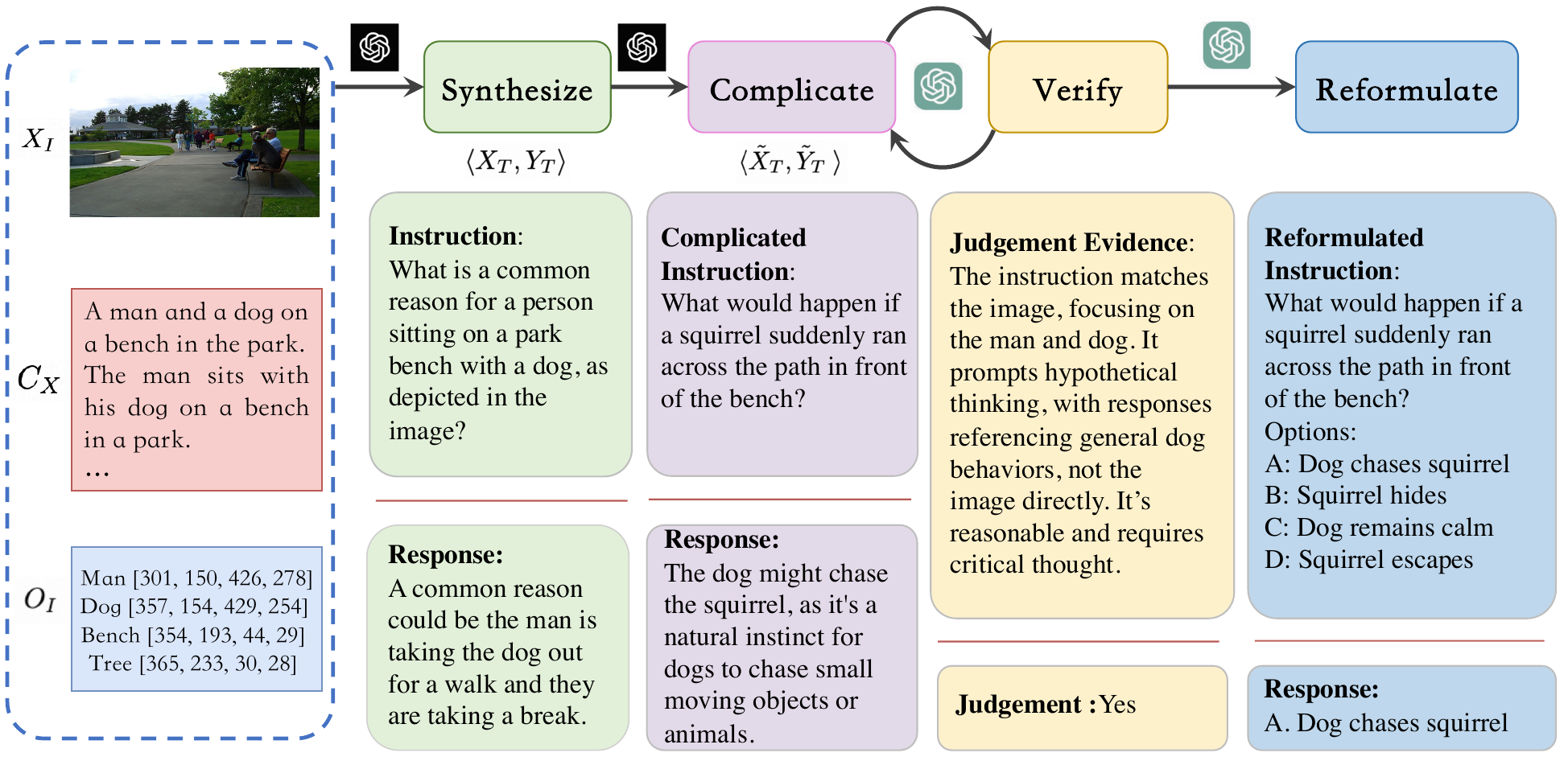}
    \caption{Our approach to synthesizing complex visual reasoning instructions involves three stages: synthesizing the primary reasoning instructions, iterative complication and verification, and reformulating instruction formats.}
    \label{fig:pipeline}
\end{figure*}
\section{Approach}
\label{sec:approach}

Based on the findings in Section~\ref{sec:empirical_results_analysis}, we propose a method to create high-quality, complex visual reasoning instructions to improve the performance of MLLMs.
For image $X_{I}$ with captions $\{C_{I}\}$ and objects $\{O_{I}\}$, we first synthesize two common kinds of visual reasoning instructions, \ie cross-modal reasoning instruction $\langle X^{(C)}_{T}, Y^{(C)}_{T} \rangle$ and outside-knowledge reasoning instruction $\langle X^{(K)}_{T}, Y^{(K)}_{T} \rangle$. Subsequently, we use an iterative complicate-then-verify procedure to gradually improve their complexity and quality, obtaining $\langle \Tilde{X}^{(C)}_{T}, \Tilde{Y}^{(C)}_{T} \rangle$ and $\langle \Tilde{X}^{(K)}_{T}, \Tilde{Y}^{(K)}_{T} \rangle$. Finally, we merge and reformulate these instructions to increase data format diversity, resulting in the final dataset.

\subsection{Visual Reasoning Instruction Synthesis}
\label{sec:reasoning_instruction_synthesis}
Cross-modal reasoning~\cite{hudson-2019-cvpr-gqa} and outside-knowledge reasoning~\cite{marino-2019-cvpr-okvqa} are key visual reasoning tasks that focus on image content and external knowledge, respectively. We synthesize instructions for both types to capture their essential information, resulting in the primary cross-modal and outside-knowledge reasoning instructions.

\subsubsection{Cross-modal Reasoning Instructions.}
\label{sec:cross_modal_reasoning_instruction}

We aim to synthesize cross-modal reasoning instruction that requires MLLMs to accurately map text entities to image objects and describe object relationships in natural language. To achieve this, we first select images that contain rich objects and then utilize GPT-4 to generate the instruction based on the image annotations.

\paragraph{Image Selection.} 
To ensure the instruction complexity, we select images containing diverse objects and relationships from the Flickr30k Entities dataset~\cite{plummer-2015-iccv-flickr30k}. Each image has five detailed captions linking objects with entities and coordinates. Empirically, we find that more informative images usually come with more detailed captions, so we consider the total character count in the five captions for each image as the indicator of informativeness. We filter out those with fewer than 700 characters in the caption and retain only the most informative ones for instruction synthesis.

\paragraph{Instruction Generation.} 
After selecting informative images and their associated captions and objects, \ie $\{\langle X_{I}, C_{I}, O_{I} \rangle\}$, we employ GPT-4 to generate cross-modal reasoning instructions $\{\langle X_{I}, X_T^{(C)}, Y_T^{(C)}\rangle\}$ as follows:
\begin{equation}
    X_T^{(C)}, Y_T^{(C)}=\text{GPT-4}(P^C, C_{I}, O_{I})
\end{equation}
where $P^C$ is the prompt for cross-modal reasoning instructions. We carefully design the prompt to instruct GPT-4 to synthesize three instructions simultaneously, while guaranteeing instruction diversity. By incorporating specific requirements and in-context demonstrations into $P^C$, we can reduce the probability of synthesizing instructions that are too simple or contain irrelevant information. The prompt is shown in Figure~\ref{fig:cm_reasoning_prompt} of the Appendix.

\subsubsection{Outside-knowledge Reasoning Instruction.}
\label{sec:outside_knowledge_reasoning_instruction}
In addition to understanding the visual semantics of an image, MLLMs also require world knowledge and common sense to help understand complex relationships in complex tasks. Following the process in Section~\ref{sec:cross_modal_reasoning_instruction}, we employ image selection and instruction generation for synthesizing outside-knowledge reasoning instructions.

\paragraph{Image Selection.} 
To synthesize outside-knowledge reasoning instructions, we require detailed object information from images~(\eg the brand of a T-shirt) to capture outside knowledge~(\eg the price of the T-shirt). We select Visual Genome~\cite{krishna-2017-ijcv-visual} as the image source, which provides an average annotation of 21 objects per image, each with a corresponding caption. However, if an image contains an excessive number of object annotations, these annotations often lack detailed information about individual objects, which is essential for the generation of outside-knowledge reasoning instructions. Hence, we set a threshold (\eg 7) and remove images exceeding this limit to ensure quality. 


\paragraph{Instruction Generation.}
To generate high-quality outside-knowledge reasoning instructions, we first select suitable objects as topic entities from the chosen images, and then prompt GPT-4 to synthesize instructions about them. For topic entity selection, we mainly consider long-tail world knowledge that MLLMs may overlook. Specifically, we utilize \emph{Inverse Document Frequency}~(IDF) to measure the importance of a certain object. We select the object with the highest IDF and denote it as $O^{'}_{I}$. Based on the topic entity and image annotation, we utilize GPT-4 to generate the instructions $\{\langle X_I, X_T^{(K)}, Y_T^{(K)} \rangle\}$ as:
\begin{equation}
    X_T^{(K)}, Y_T^{(K)}=\text{GPT-4}(P^K, C_{I}, O^{'}_{I})
\end{equation}
where $P^K$ is the prompt for outside-knowledge instructions. Additionally, we sample knowledge categories from the category set in OK-VQA~\cite{marino-2019-cvpr-okvqa} and incorporate them into $P^K$, guiding GPT-4 to produce instructions related to these categories. This ensures balanced knowledge coverage across the instruction dataset. The detailed prompt is shown in Figure~\ref{fig:ok_reasoning_prompt} of the Appendix.

\subsection{Visual Reasoning Instruction Complication}
\label{sec:visual_instruction_improvement}
Through this instruction synthesis process, we obtain two types of instruction sets. Despite the carefully designed prompts, the synthetic instructions are still relatively simple and even contain hallucinated objects. To address this, we propose an iterative \textit{complicate-then-verify} procedure to gradually increase the complexity of the instructions and meanwhile ensure the quality and avoid contradictions or hallucinations.

\paragraph{Instruction Complication.}
Inspired by existing work~\cite{xu-2023-arxiv-wizardlm}, we instruct GPT-4 to iteratively complicate the instructions and generate the corresponding response, based on the primary  instructions and image annotations, denoted as:
\begin{equation}
    \Tilde{X}_T, \Tilde{Y}_T=\text{GPT-4}(P^{Comp}, X_T, Y_T, C_{I}, O_{I})
\end{equation}
where $P^{Comp}$ is the prompt sent to GPT-4, shown in Figure~\ref{fig:complicate_prompt} in the Appendix. Empirically, we find that very few iteration turns (\eg 1 or 2) are sufficient to obtain high-quality instructions, thereby reducing the cost of APIs invocation.

\paragraph{Instruction Verification.}
To ensure instruction quality, we use a verification process to filter out instructions that contradict the image. Specifically, we prompt ChatGPT to determine if the synthesized instruction aligns with the provided image annotations. The prompt is shown in Figure~\ref{fig:verify_prompt} in the Appendix. Based on the judgment of ChatGPT, we only retain the instructions that pass the verification and discard the failed ones.

\subsection{Visual Reasoning Instruction Reformulation}
\label{sec:visual_instruction_reformulation}
After the synthesis and complication processes, we obtain many high-quality, complex instructions. However, these open-ended responses may not suit tasks requiring specific formats (\eg multiple-choice or boolean QA), potentially affecting zero-shot generalization.

To address this, we incorporate a reformulation stage, in which we sample some synthetic instructions and use ChatGPT to convert them into two distinct representative formats: boolean QA and multiple-choice QA. Boolean QA offers binary answers, \ie ``yes'' or ``no'', while multiple-choice QA provides several predefined options. After the reformulation stage, we combine the original open-ended instructions with the newly reformulated instructions to create the final \textbf{ComVint} dataset.

\begin{table*}[t]
    \centering
    \scalebox{0.8}{
    \begin{tabular}{cc|cccccccc}
    \toprule
         Model&  Benchmark&  Original&  \makecell[c]{+LLaVA\\Cap}&  \makecell[c]{+LLaVA\\VQA}&  \makecell[c]{+LLaVA\\Rea}&  +LRV&  +\makecell[c]{Shikra\\-RD}& +A-OKVQA
  & +ComVint\\
    \midrule
         \multirow{3}{*}{MiniGPT4}&  SEED-Bench&  43.31&  \cellcolor[HTML]{DDEBF4}41.60&  \cellcolor[HTML]{FEF1EA}45.97&  \cellcolor[HTML]{FEF1EA}43.69&  \cellcolor[HTML]{FC8D59}\textbf{50.92}&  \cellcolor[HTML]{DDEBF4}41.75& \cellcolor[HTML]{FEF1EA}43.99
  & \cellcolor[HTML]{FC8D59}\underline{50.13}\\
         &  MME-P&  820.37&  \cellcolor[HTML]{A7CBE2}643.59&  \cellcolor[HTML]{A7CBE2}650.51&  \cellcolor[HTML]{FC8D59}\underline{929.82}&  \cellcolor[HTML]{A7CBE2}552.75&  \cellcolor[HTML]{C4DCEC}722.38& \cellcolor[HTML]{FC8D59}924.47
  &\cellcolor[HTML]{FEF1EA}856.30 \\
         &  MME-C&  198.57&  \cellcolor[HTML]{DDEBF4}154.29&  \cellcolor[HTML]{FEF1EA}218.21&  \cellcolor[HTML]{FDD1BD}\underline{265.71}&  \cellcolor[HTML]{FEF1EA}199.29&  \cellcolor[HTML]{DDEBF4}172.14& \cellcolor[HTML]{FDD1BD}257.50
  & \cellcolor[HTML]{FEF1EA}227.14\\
 
        \midrule
         \multirow{3}{*}{BLIP-2}&  SEED-Bench&  53.68&  \cellcolor[HTML]{DDEBF4}52.31 &  \cellcolor[HTML]{C4DCEC}51.00&  \cellcolor[HTML]{C4DCEC}50.94 &  \cellcolor[HTML]{FEF1EA}\textbf{54.64} &  \cellcolor[HTML]{DDEBF4}52.92 & \cellcolor[HTML]{DDEBF4}52.60 
  & \cellcolor[HTML]{FEF1EA}\underline{53.73} \\
         &  MME-P&  1151.26  &  \cellcolor[HTML]{DDEBF4}1115.52  &  \cellcolor[HTML]{C4DCEC}907.21  &  \cellcolor[HTML]{FEF1EA}\underline{1162.47}  &  \cellcolor[HTML]{A7CBE2}643.02  &  \cellcolor[HTML]{C4DCEC}1102.43 & \cellcolor[HTML]{DDEBF4}1115.69  
  & \cellcolor[HTML]{FDD1BD}\textbf{1216.24} \\
         &  MME-C&  241.07  &  \cellcolor[HTML]{DDEBF4}224.64  &  \cellcolor[HTML]{DDEBF4}238.93  &  \cellcolor[HTML]{FEF1EA}\textbf{271.07}  &  \cellcolor[HTML]{DDEBF4}216.43  &  \cellcolor[HTML]{DDEBF4}231.43 & \cellcolor[HTML]{FEF1EA}247.14  
  & \cellcolor[HTML]{FEF1EA}\underline{250.71} \\
 
        \midrule
 \multirow{3}{*}{LLaVA}& SEED-Bench& 49.43& \cellcolor[HTML]{C4DCEC}48.52& \cellcolor[HTML]{92BFDB}46.86& \cellcolor[HTML]{92BFDB}46.82& \cellcolor[HTML]{FC8D59}\textbf{56.58}& \cellcolor[HTML]{DDEBF4}49.01&\cellcolor[HTML]{FDD1BD}54.01
  & \cellcolor[HTML]{FDD1BD}\underline{54.74}\\
 & MME-P& 949.42& \cellcolor[HTML]{FDD1BD}1091.41& \cellcolor[HTML]{FDD1BD}1002.42& \cellcolor[HTML]{FDD1BD}1043.53& \cellcolor[HTML]{FC8D59}\underline{1154.99}& \cellcolor[HTML]{C4DCEC}915.64&\cellcolor[HTML]{FC8D59}1140.23
  & \cellcolor[HTML]{FC8D59}\textbf{1213.87}\\
 & MME-C& 232.86& \cellcolor[HTML]{FEF1EA}238.21& \cellcolor[HTML]{FEDCCC}257.14& \cellcolor[HTML]{C4DCEC}211.07& \cellcolor[HTML]{FDD1BD}272.14& \cellcolor[HTML]{FDD1BD}260.00&\cellcolor[HTML]{FC8D59}\textbf{310.36}  & \cellcolor[HTML]{FC8D59}\underline{297.14}\\

        \midrule
         \multirow{3}{*}{InstructBLIP}&  SEED-Bench&  56.14&  \cellcolor[HTML]{DDEBF4}55.44 &  \cellcolor[HTML]{C4DCEC}52.63 &  \cellcolor[HTML]{DDEBF4}55.22 &  \cellcolor[HTML]{FEF1EA}56.57 &  \cellcolor[HTML]{FEF1EA}\textbf{57.49} & \cellcolor[HTML]{DDEBF4}56.12 
  & \cellcolor[HTML]{FEF1EA}\textbf{57.49} \\
         &  MME-P&  1178.95  &  \cellcolor[HTML]{FEF1EA}\underline{1211.72}  &  \cellcolor[HTML]{A7CBE2}972.29  &  \cellcolor[HTML]{FEF1EA}1205.13  &  \cellcolor[HTML]{A7CBE2}918.15  &  \cellcolor[HTML]{DDEBF4}1176.63  & \cellcolor[HTML]{FC8D59}\textbf{1262.29}  & \cellcolor[HTML]{FEF1EA}1199.91 \\
         &  MME-C&  301.79 &  \cellcolor[HTML]{FDD1BD}\textbf{340.00}  &  \cellcolor[HTML]{FEF1EA}302.86  &  \cellcolor[HTML]{DDEBF4}287.14  &  \cellcolor[HTML]{C4DCEC}219.64  &  \cellcolor[HTML]{DDEBF4}279.29  & \cellcolor[HTML]{DDEBF4}277.50  
  & \cellcolor[HTML]{FEF1EA}\underline{305.36} \\
 \bottomrule
    \end{tabular}}
    \caption{The performance of four representative MLLMs fine-tuned on different instruction collections\ignore{, where ``MME-P'' and ``MME-C'' represent MME-Perception and MME-Cognition, respectively}. The best performance and the second-best performance are denoted in bold and underlined fonts, respectively. \ignore{Cells shaded in orange indicate that fine-tuning enhances the original MLLM performance, and blue indicates performance degradation.}}
    \label{tab:main_result}
\end{table*}

To evaluate the quality of the data synthesized by GPT-4, we randomly sample some instances for human review. The results in Table~\ref{tab:manual_analysis} in the Appendix show that most instructions are of high quality. We also compare cases in LLaVA-Reasoning, LRV, and our ComVint in Figure~\ref{fig:case_study} in the Appendix, and find that instructions in ComVint are more complex and involve more reasoning steps.

\section{Experiment}
\label{sec:experiments}

\subsection{Experimental Setup}
To exhibit the generality of our instructions, we fine-tune four representative MLLMs on ComVint: BLIP-2, MiniGPT-4, LLaVA, and InstructBLIP. These models were selected because they employ diverse architectures, training strategies, and training datasets, making them ideal for verifying the generalizability of our conclusions. A detailed introduction to these models is in the Appendix. We fine-tune these models on our instruction dataset and the other six representative visual instruction collections~(LLaVA Caption, LLaVA VQA, LLaVA Reasoning, LRV, Shikra-RD, and A-OKVQA) used in Section~\ref{sec:empirical} for comparison.

\paragraph{Implementation Details.}
For BLIP-2, MiniGPT-4, and InstructBLIP, we set the learning rate to 1e-5 and train for 2 epochs, while for LLaVA, we set the learning rate to 2e-5 and train for 3 epochs. The batch size for BLIP-2, LLaVA, and InstructBLIP is 128, while the batch size for MiniGPT-4 is 64. Concerning the mixture of our synthesized instructions, the final instruction collection comprises approximately 12K cross-modal reasoning instructions and 20K outside-knowledge reasoning instructions. 

\paragraph{Evaluation Benchmarks.}
We follow Section~\ref{sec:empirical} and mainly evaluate the models on SEED-Bench Image and MME. Additionally, we also incorporate three traditional visual reasoning benchmarks, \ie OK-VQA~\cite{marino-2019-cvpr-okvqa}, A-OKVQA~\cite{schwenk-2022-eccv-aokvqa}, and GQA~\cite{hudson-2019-cvpr-gqa}. \ignore{The questions in OK-VQA and A-OKVQA span diverse knowledge categories, and the model needs to associate world knowledge with the objects in the image to answer the questions. GQA mainly focuses on cross-modal reasoning questions, requiring the model to perform reasoning within an image.}

\subsection{Main Results}

The results of the four models fine-tuned on seven instruction collections are shown in Table~\ref{tab:main_result}. Based on the results, we have the following findings:

\ignore{Firstly, while LLaVA-Rea and A-OKVQA are the most effective instruction sets among the previous ones, simply combining them does not always yield the best performance and even results in performance degradation on most models. This is possibly due to the fact that the instructions in these two datasets have a large gap and thus conflict with each other. In comparison, ComVint consistently enhances model strengths and improves their relatively weaker capabilities, leading to consistent improvements across all benchmarks.}

Firstly, among all the existing visual instruction collections, complex visual reasoning instructions~(\ie A-OKVQA and LLaVA Reasoning) generally lead to the most substantial improvements across the three evaluation dimensions compared to others. The enhancements in InstructBLIP and LLaVA are not as significant, as they had already utilized these data during instruction tuning.

Secondly, baseline instruction datasets can enhance model performance on specific benchmarks. For instance, MiniGPT-4, BLIP-2, and LLaVA fine-tuned on LRV outperform those fine-tuned on ComVint when evaluated on the SEED-Bench benchmark. This is because LRV is specially designed with valuable features, such as various task types and formats, which can be beneficial for improving MLLMs in related tasks. However, introducing certain characteristics can degrade performance on unrelated tasks. For example, BLIP-2 fine-tuned on LRV shows significant drops in MME benchmark~(MME-Perception decreases from 1151.26 to 643.02, and MME-Cognition decreases from 241.07 to 216.43). In comparison, our ComVint effectively balances various capabilities and yields improvements across all benchmarks. We also display the results of LLaVA on all the sub-tasks of SEED-Bench in Table~\ref{tab:appendix_subtask} in the Appendix. We can observe that ComVint significantly improves the performance on all the sub-tasks.

\begin{table}[t]
\small
  \centering
    \begin{tabular}{lccc}
    \toprule
         {Instruction}&{OK-VQA}&A-OKVQA&{GQA}\\
    \midrule
         Original&57.82& 77.12& 50.98\\
         +\textcircled{1}: LLaVA-Rea&   46.54& 72.93& 46.88\\
         +\textcircled{2}: A-OKVQA&   57.48& \textcolor{gray}{78.86*}& 50.90\\
         +\textcircled{3}: ComVint& \textbf{58.71}&\textbf{78.08}& \textbf{51.75}\\
    \bottomrule
    \end{tabular}
    \caption{Results of InstructBLIP on traditional VQA evaluation benchmarks. The result marked with * denotes that the model is fine-tuned on the training set of the evaluation benchmark.}
    \label{tab:more_benchmark}
\end{table}

Thirdly, since InstructBLIP achieves the best performance on these benchmarks, we also evaluate it on three traditional VQA benchmarks: OK-VQA~\cite{marino-2019-cvpr-okvqa}, A-OKVQA~\cite{schwenk-2022-eccv-aokvqa}, and GQA~\cite{hudson-2019-cvpr-gqa}. The results in Table~\ref{tab:more_benchmark} show that all previous instruction datasets hurt performance on these benchmarks, while ComVint is the only instruction dataset that can further boost performance.

\begin{figure*}[t]
    \centering
    \includegraphics[width=0.9\linewidth]{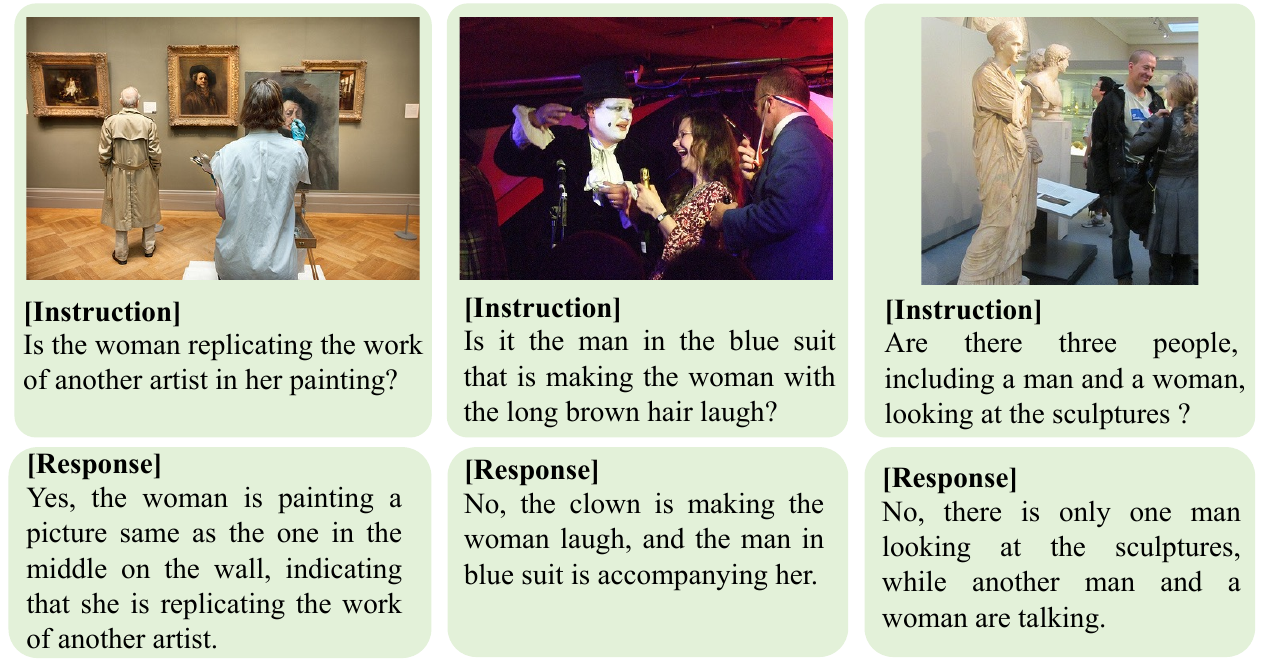}
    \caption{Examples from the ComVint dataset, highlighting instructions designed to emphasize cross-modal reasoning over basic visual perception.}
    \label{fig:appendix_case}
\end{figure*}

Besides the quantitative results, we present case examples from ComVint in Figure~\ref{fig:appendix_case} and Figure~\ref{fig:case_study}, comparing them to previous instruction sets. We can observe that LLaVA-Reasoning primarily focuses on scene descriptions, while LRV emphasizes object recognition. In contrast, ComVint is more complex and includes more reasoning steps.


\subsection{In-Depth Analysis}
\label{sec:ablation}
\paragraph{Data Quality Analysis.} We randomly sample 100 instances from ComVint, LRV, and LLaVA-Reasoning and have three of our authors manually assess their quality. Following prior work~\cite{liu-2023-arxiv-aligning}, we separately check whether the instructions and responses are correct. The evaluation criteria for correctness are in the Appendix. We calculate the Fleiss' Kappa among the three annotators, obtaining a value of 0.91, which indicates near-perfect agreement. As shown in Table~\ref{tab:manual_analysis}, most instructions are accurate. Besides, the responses in ComVint exhibit comparable quality to LLaVA-Reasoning, and superior to LRV, highlighting the high quality of our dataset.
\begin{table}[t]
  \centering
    \begin{tabular}{lccc}
    \toprule
            &  LRV&  LLaVA-Rea & ComVint\\
    \midrule
             Instruction &88.00  &90.00  &88.00\\
             Response&65.00 &86.00 &84.00\\
    \bottomrule
    \end{tabular}
    \caption{The correct rate of each instruction dataset evaluated by humans.}
    \label{tab:manual_analysis}
\end{table}

\paragraph{The Effect of Complication.} For each instruction set, we ask gpt-3.5-turbo-0125 to count the reasoning steps in each instruction, defining the average reasoning steps as the complexity score of the dataset. We then plot the relationship between complexity scores and the average accuracy of four models on these benchmarks. The result in Figure~\ref{fig:complexity_score} shows that as the complexity of the visual instruction datasets increases, the average performance consistently improves, following a roughly linear trend. Notably, ComVint, with the most complex instruction set, achieves the best results. Please refer to the Appendix~\ref{sec:discussion} for more details. Meanwhile, to test the effectiveness of the complication operation in our pipeline, we create a basic instruction set by removing all complexity constraints and skipping the complication stage (denoted as w/o Comp in Table~\ref{tab:ablation1}). We fine-tune LLaVA on ComVint and this basic set. The results in Table~\ref{tab:ablation1} show significant degradation on all benchmarks, demonstrating the importance of instruction complexity.

\paragraph{The Effect of Two Types of Instructions.}
In our reasoning instruction synthesis stage, we generate cross-modal and outside-knowledge reasoning instructions. To assess their effectiveness, we remove either of them and fine-tune LLaVA on the remaining instructions~(denoted as w/o O-K Rea and w/o C-M Rea in Table~\ref{tab:ablation1}). The results show that removing either type hurts performance on SEED-Bench and MME-Perception. As for the MME-Cognition tasks, removing cross-modal reasoning instructions yields the best performance. This is because answering questions in MME-Cognition most rely on cognition ability, thus our outside knowledge reasoning instructions are more beneficial. Generally, incorporating both types of instruction leads to the best average performance.

\setlength{\tabcolsep}{2.5pt} 
\begin{table}
\centering
\small
    \begin{tabular}{lcccc}
        \toprule
         Instruction&  SEED-Bench&  MME-P& MME-C &Avg\\
         \midrule
         Original&  49.43 &  39.36 & 12.31 &33.70 \\
 +ComVint& \textbf{54.74}& \textbf{55.53} & 26.15 &\textbf{45.47} \\
 \quad w/o Comp& 50.89& 44.94& 12.31&36.05\\
         \quad w/o O-K Rea&  51.40 &  49.29 & 24.62 &41.77 \\
         \quad w/o C-M Rea&  53.85 &  46.74 & \textbf{27.69} &42.76 \\
         \quad w/o Reform&  53.75&  45.60& 19.23&39.53 \\
         \bottomrule
    \end{tabular}
    \caption{Results of ablation study. ``Comp'' stands for complication, ``C-M Rea'' for cross-modal reasoning, ``O-K Rea'' for outside-knowledge reasoning, and ``Reform'' for reformulation.}
    \label{tab:ablation1}
\centering
\end{table}

\paragraph{The Effect of Instruction Format.}
\ignore{Our approach employs a reformulation module to diversify instruction formats, and we evaluate its impact on MLLM performance. Specifically, }We remove the reformulation module to obtain 32K open-ended instructions~(denoted as w/o Reform) and fine-tune LLaVA on them, with the results presented in Table~\ref{tab:ablation1}. We can observe that removing the reformulation module would lead to performance degradation compared to the instructions with diverse formats, highlighting the importance of the reformulation module. On the other hand, fine-tuning the model only on the open-ended instructions also significantly improves the performance, implying that the reformulation module is not the only factor contributing to the improvement.

\subsection{Case Study}
In this section, we display some instructions in our ComVint dataset in Figure~\ref{fig:appendix_case}. The first instruction requires the model to compare the painting of the woman and the painting on the wall, and then understand that the woman is replicating the painting on the wall. The second instruction requires the model to understand the relationships among these people. The third instruction requires the model to observe the image carefully and understand the activity of these people. We also display several cases from LLaVA-Reasoning, LRV, and ComVint in Figure~\ref{fig:case_study} for comparison. Please refer to Appendix~\ref{app:case} for more details.

\section{Conclusion}
In this paper, we investigated key factors contributing to effective visual instructions for MLLMs.\ignore{an essential question of multi-modal large language models~(MLLMs): ``\emph{What makes for good visual instructions?}''.} Our empirical experiments demonstrated that instructions with the visual reasoning task type and complexity characteristic were more useful for improving the capabilities of MLLMs. Based on these insights, we devised a systematic approach to automatically create high-quality complex visual reasoning instructions, employing a synthesize-complicate-reformulate paradigm to gradually improve the complexity while guaranteeing quality. Using this approach, we synthesized a visual reasoning instruction dataset, namely \textbf{ComVint}, for fine-tuning MLLMs. Experimental results have demonstrated the efficacy of our dataset in improving the capability of representative MLLMs.

\ignore{In future work, we will investigate to craft more high-quality visual instructions in an automatic way or with limited human annotation. Furthermore, we also consider improving the performance of MLLMs from the perspective of training strategy.}
\section{Limitation}
First, many factors influence the final performance of an MLLM: training data, model architecture, and training strategies, \etc. In this work, we focus solely on training data, identifying what makes for good visual instructions regarding task type and instruction characteristics, while leaving other factors unchanged. In the future, we plan to explore how these factors, in coordination with training data, affect the overall performance of MLLMs. Second, the scale of ComVint is not large enough. While we achieve strong results with limited data, demonstrating the value of ``less is more'', we believe that scaling up the amount of high-quality visual instruction data will further enhance performance.
\section*{Acknowledgements}
This work was partially supported by National Natural Science Foundation of China under Grant No. 92470205 and 62222215. This research was also supported by Meituan and the Outstanding Innovative Talents Cultivation Funded Programs 2023 of Renmin University of China.
Xin Zhao is the corresponding author. 
\bibliography{custom}

\appendix

\clearpage
\setcounter{page}{1}

\appendix
\label{sec:appendix}

\section{Case Study}
\label{app:case}
\begin{figure}[h]
    \centering
    \includegraphics[width=1\linewidth]{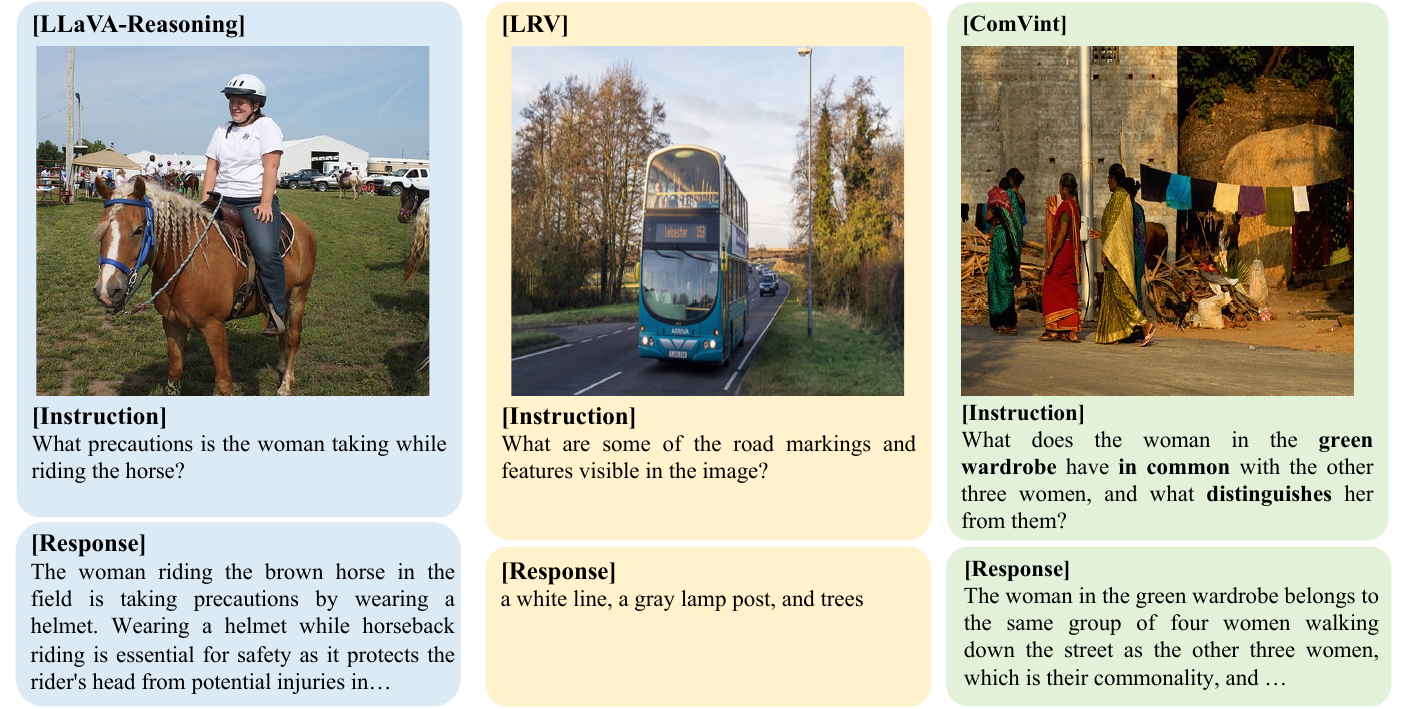}
    \caption{Sampled instructions from LLaVA-Reasoning, LRV, and ComVint, where the instruction in ComVint is more complex than LLaVA-Reasoning and LRV.}
    \label{fig:case_study}
\end{figure}

We also display several cases from LLaVA-Reasoning, LRV, and ComVint in Figure~\ref{fig:case_study} for comparison. The instruction in LLaVA-Reasoning mainly focuses on the description of the scene, which can be answered easily once the model recognizes the helmet in the image. The instruction in LRV is relatively shorter and simpler, focusing on the object recognition ability of the model. In contrast, the instruction in ComVint is more complex and includes more reasoning steps. To answer the question in the image, the model needs to first identify the three women according to the color of their clothes. Then, the model needs to identify their actions and finally, distinguish the commonalities and differences between them.

\section{Baseline Models}
In this section, we elaborate on the details of our baseline models.
\begin{itemize}[left=0pt]
    \item \textit{BLIP-2}~\cite{Li-2023-ICML-BLIP2} is based on FLAN-T5-XXL and utilizes a Q-Former to connect the vision encoder and the LLM. It freezes the vision encoder and the LLM during fine-tuning and only updates the parameters of the Q-Former. The training data contains 129M image-caption pairs.

    \item \textit{MiniGPT-4}~\cite{zhu-2023-arxiv-MiniGPT} is based on Vicuna-7B and reuses the Q-Former in BLIP-2 to connect the vision encoder and the LLM. It freezes the Q-Former, the vision encoder, and the LLM during fine-tuning and only updates the parameters of the linear layer between the Q-Former and the LLM. The training data includes 5M image-text pairs and 3,500 high-quality image-caption pairs.

    \item \textit{LLaVA}~\cite{liu-2023-arxiv-visual} is based on Vicuna-7B and utilizes linear layers to connect the vision encoder and the LLM. It freezes the vision encoder during fine-tuning and updates the linear layers and the LLM. The training data is 158K visual instructions, including 58K in conversations, 23K in detailed descriptions, and 77K in visual reasoning.

    \item \textit{InstructBLIP}~\cite{dai-2023-arxiv-instructblip} is based on FLAN-T5-XXL and the architecture and the trainable parameters are the same as BLIP-2. The training data includes 10 vision-language tasks as well as the instruction data used in LLaVA.
\end{itemize}

\section{Discussion}
\label{sec:discussion}
To analyze the impact of instruction complexity on the performance of the MLLM, we randomly sample 1000 instructions from each instruction set listed in Table~\ref{tab:main_result}. Utilizing the gpt-3.5-turbo-0125 model, we parse these instructions into a series of sub-questions, with each sub-question representing a reasoning step. The complexity of a question is measured by the number of reasoning steps, and we calculate the average number of reasoning steps for each instruction set to derive its complexity score. Subsequently, we average the results for the MME and SEED-Bench benchmarks as presented in Table~\ref{tab:main_result}. Further, we average the results across the four MLLMs to obtain the final average accuracy. The results depicted in Table~\ref{fig:complexity_score} reveal that, overall, an increase in instruction complexity corresponds to an improvement in MLLM performance, demonstrating a trend that can be approximated by a linear function. An exception is observed with LRV, possibly due to its inclusion of more incorrect instructions, as indicated in Table~\ref{tab:manual_analysis}. Moreover, LLaVA-Cap, characterized by the lowest complexity, yields the lowest accuracy, while LLaVA-Rea and A-OKVQA, with higher complexity, exhibit better performance, aligning with findings from our empirical study in Table~\ref{sec:empirical}. Notably, ComVint, with the most complex instruction set, achieves the best results.

\begin{table*}[h]
    \centering
    \scalebox{0.85}{
    \begin{tabular}{lcccccccccc}
    \toprule
 Inst & \makecell[c]{instance\\counting}& \makecell[c]{instance\\attributes}& \makecell[c]{scene\\understanding}& \makecell[c]{instance\\identity}& \makecell[c]{instance\\interaction}& \makecell[c]{visual\\reasoning}& \makecell[c]{instance\\location}& \makecell[c]{spatial\\relations}& \makecell[c]{text\\recognition}&Average\\
 \midrule
 Original& 38.95& 51.90& 57.09& 53.69& 47.42& 47.43& 41.21& 38.36& 30.59&49.43\\
         +ComVint&  \textbf{42.54}&  \textbf{59.65}&  \textbf{62.13}&  \textbf{56.96}&  \textbf{57.73}&  \textbf{53.47}&  \textbf{44.38}&  \textbf{40.79}&  \textbf{43.53} &\textbf{54.74}\\
    \bottomrule
    \end{tabular}}
    \caption{The performance of LLaVA on all the sub-tasks of SEED-Bench Image.}
    \label{tab:appendix_subtask}
\end{table*}

\section{More Experimental Results}
We display the accuracy of LLaVA on all the sub-tasks on the SEED-Bench Image in Table~\ref{tab:appendix_subtask}. The results show that ComVint can improve the performance of LLaVA on all the sub-tasks on SEED-Bench Image, especially on instance attributes, instance interaction, visual reasoning, and text recognition. This demonstrates the effectiveness of our instruction dataset.

\section{Prompt Design}

\begin{figure*}
    \centering
    \includegraphics[width=0.9\linewidth]{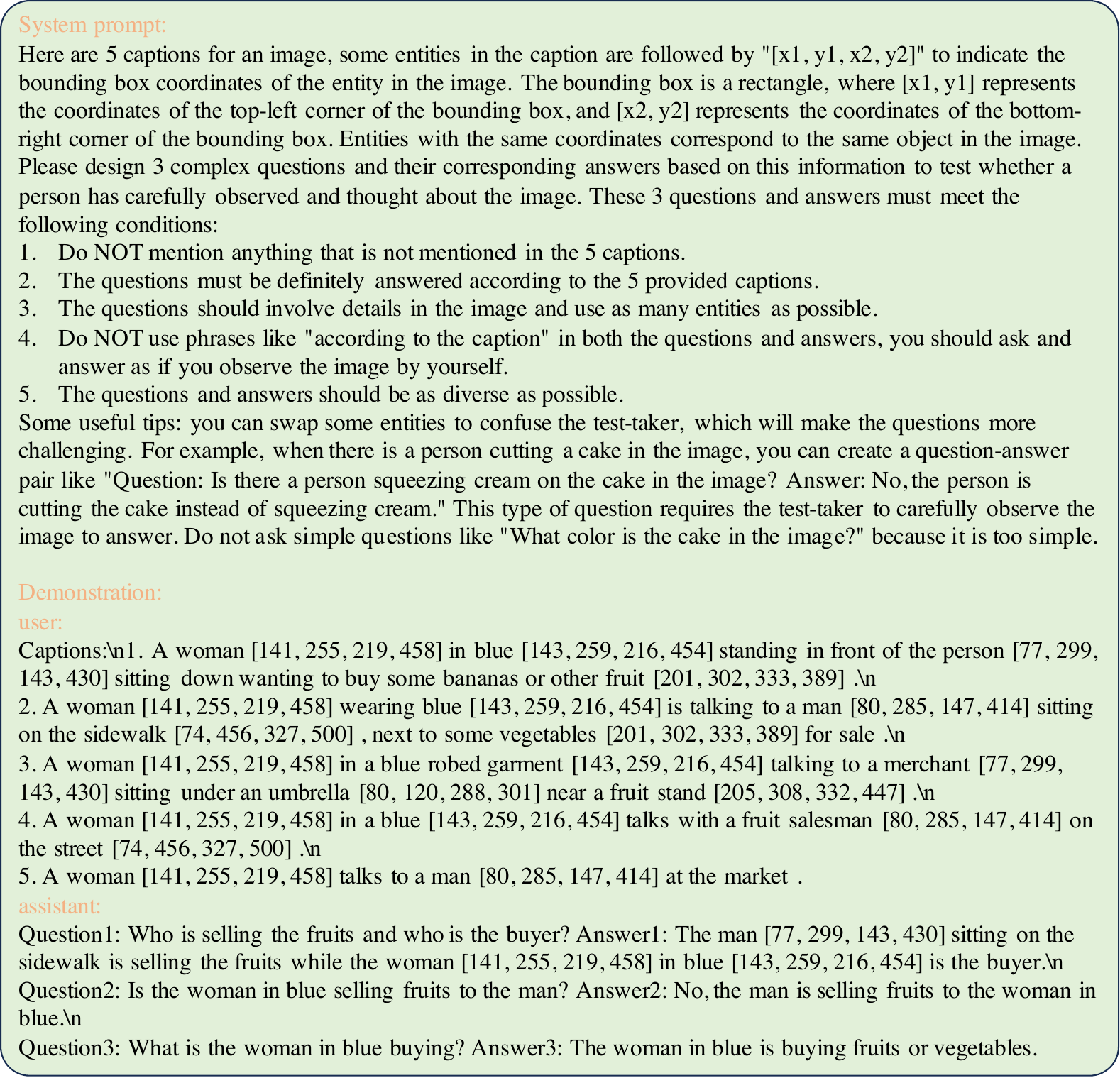}
    \caption{The prompt we give to GPT-4 for constructing cross-modal reasoning instructions.}
    \label{fig:cm_reasoning_prompt}
\end{figure*}

\begin{figure*}
    \centering
    \includegraphics[width=0.9\linewidth]{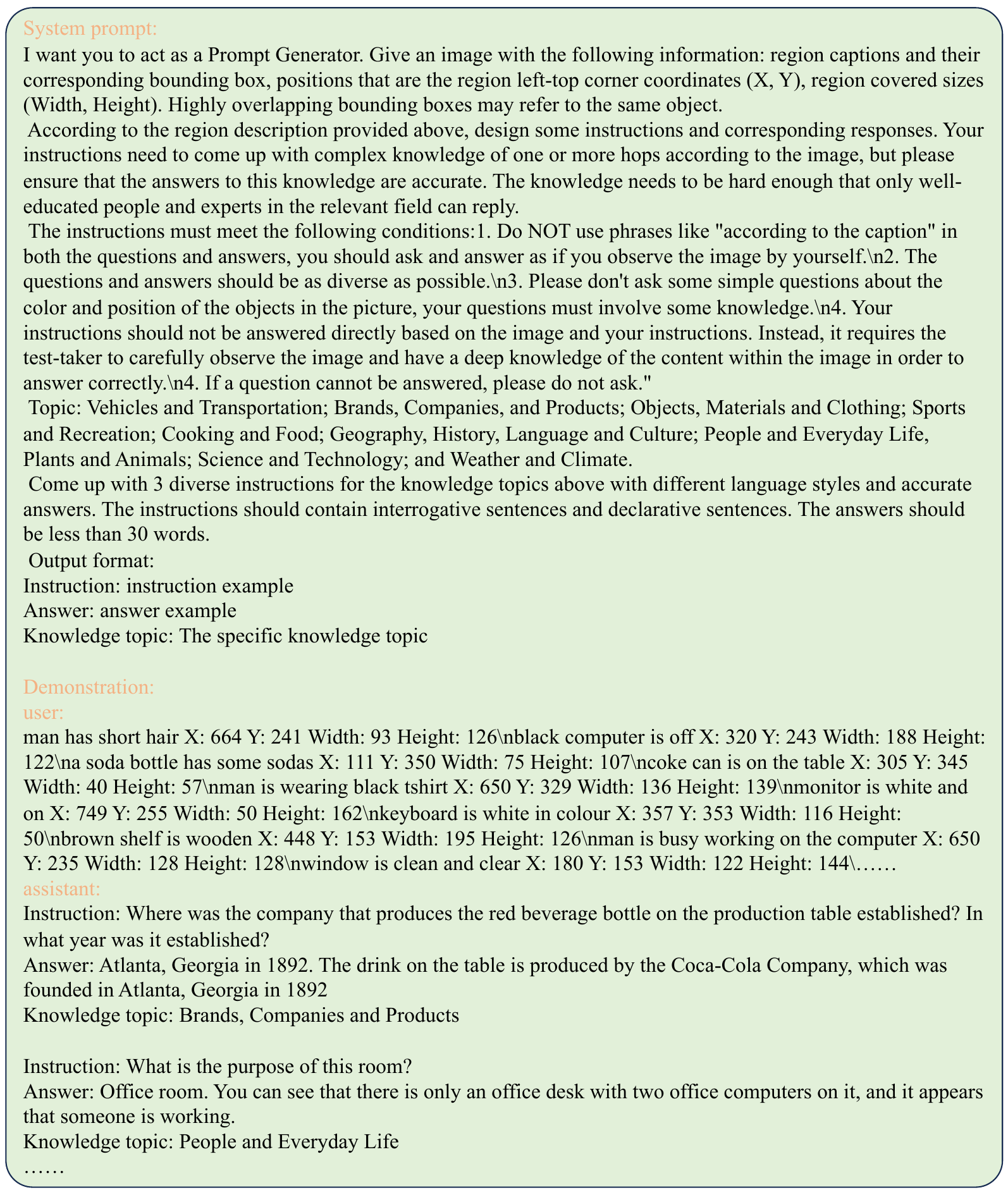}
    \caption{The prompt we give to GPT-4 for constructing outside-knowledge reasoning instructions.}
    \label{fig:ok_reasoning_prompt}
\end{figure*}

\begin{figure*}[htbp]
    \centering
    \includegraphics[width=0.9\linewidth]{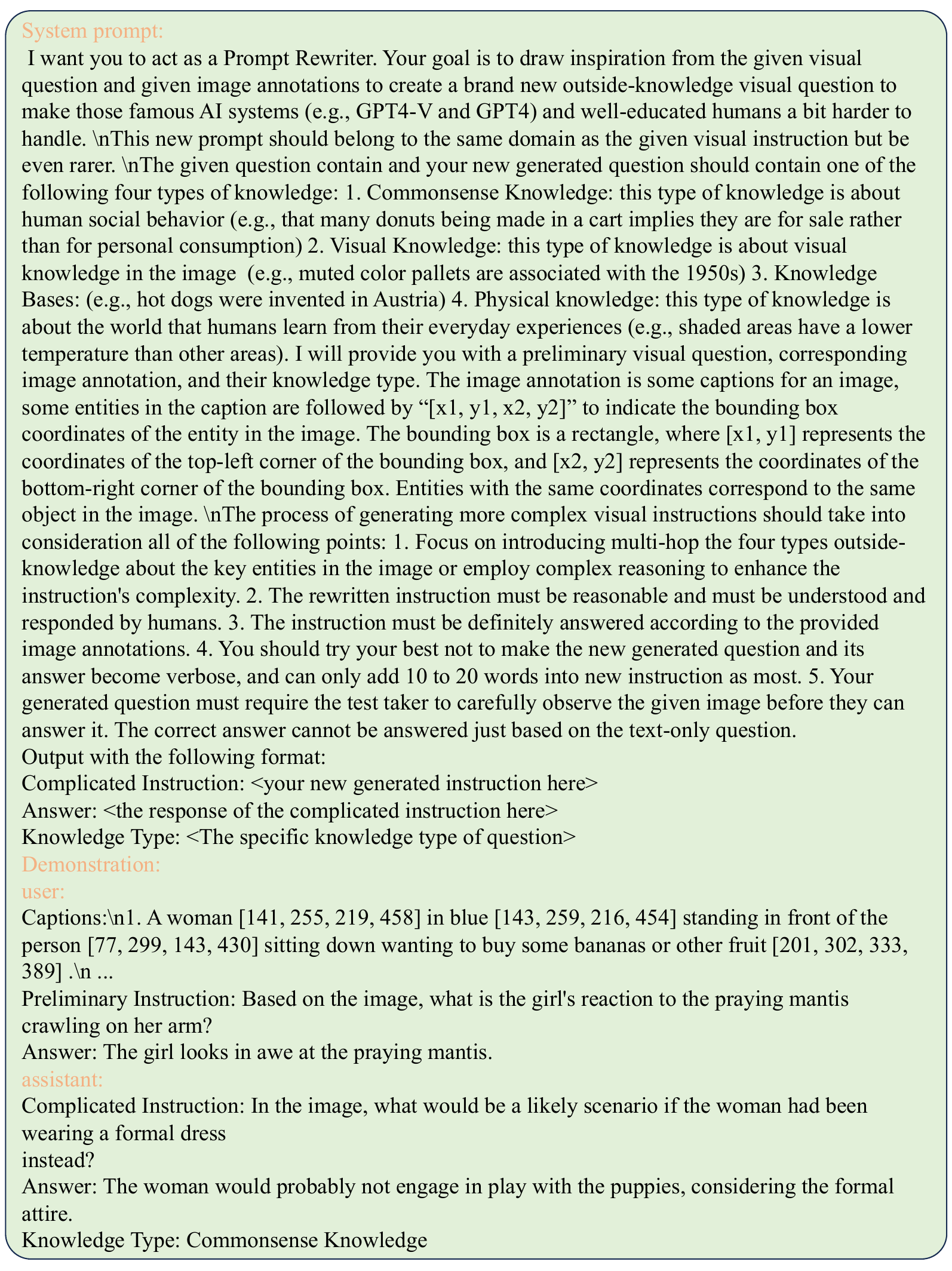}
    \caption{The prompt we give to GPT-4 for complicating instructions.}
    \label{fig:complicate_prompt}
\end{figure*}

\begin{figure*}[htbp]
    \centering
    \includegraphics[width=0.9\linewidth]{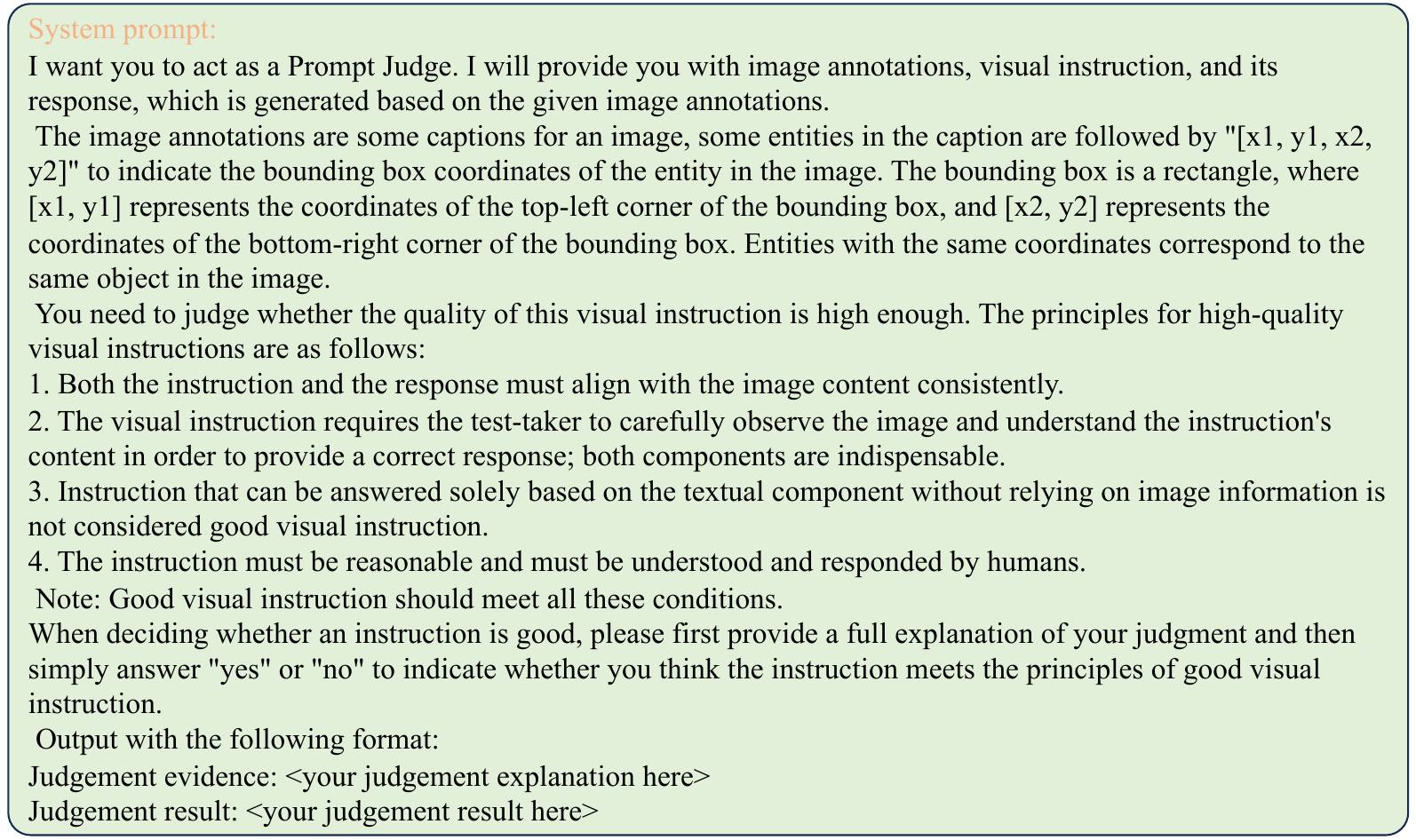}
    \caption{The prompt we give ChatGPT for verifying instructions.}
    \label{fig:verify_prompt}
\end{figure*}

\begin{figure*}[htbp]
    \centering
    \includegraphics[width=0.9\linewidth]{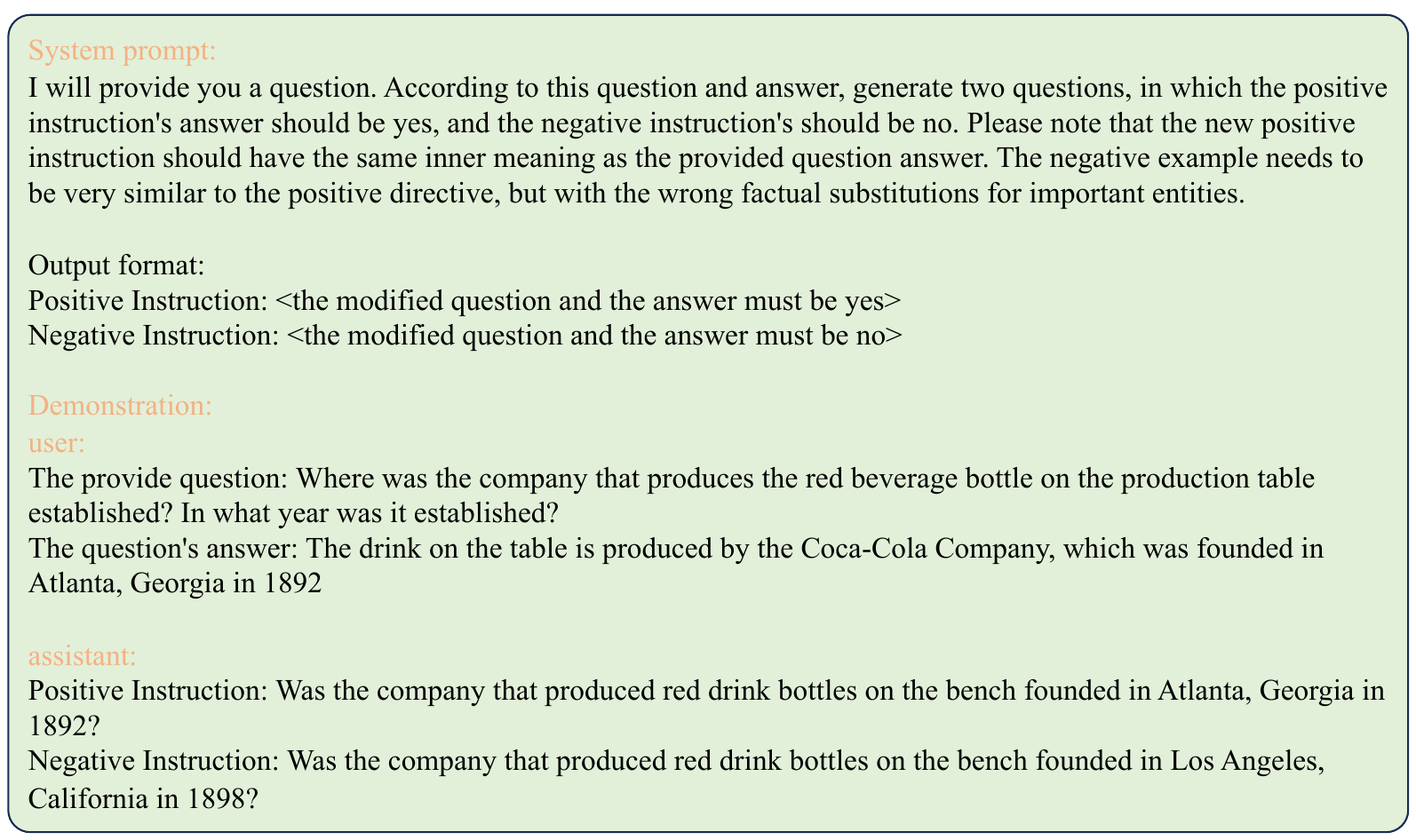}
    \caption{The prompt we give ChatGPT for generating Bool QA instructions.}
    \label{fig:boolqa_prompt}
\end{figure*}

\begin{figure*}[htbp]
    \centering
    \includegraphics[width=0.9\linewidth]{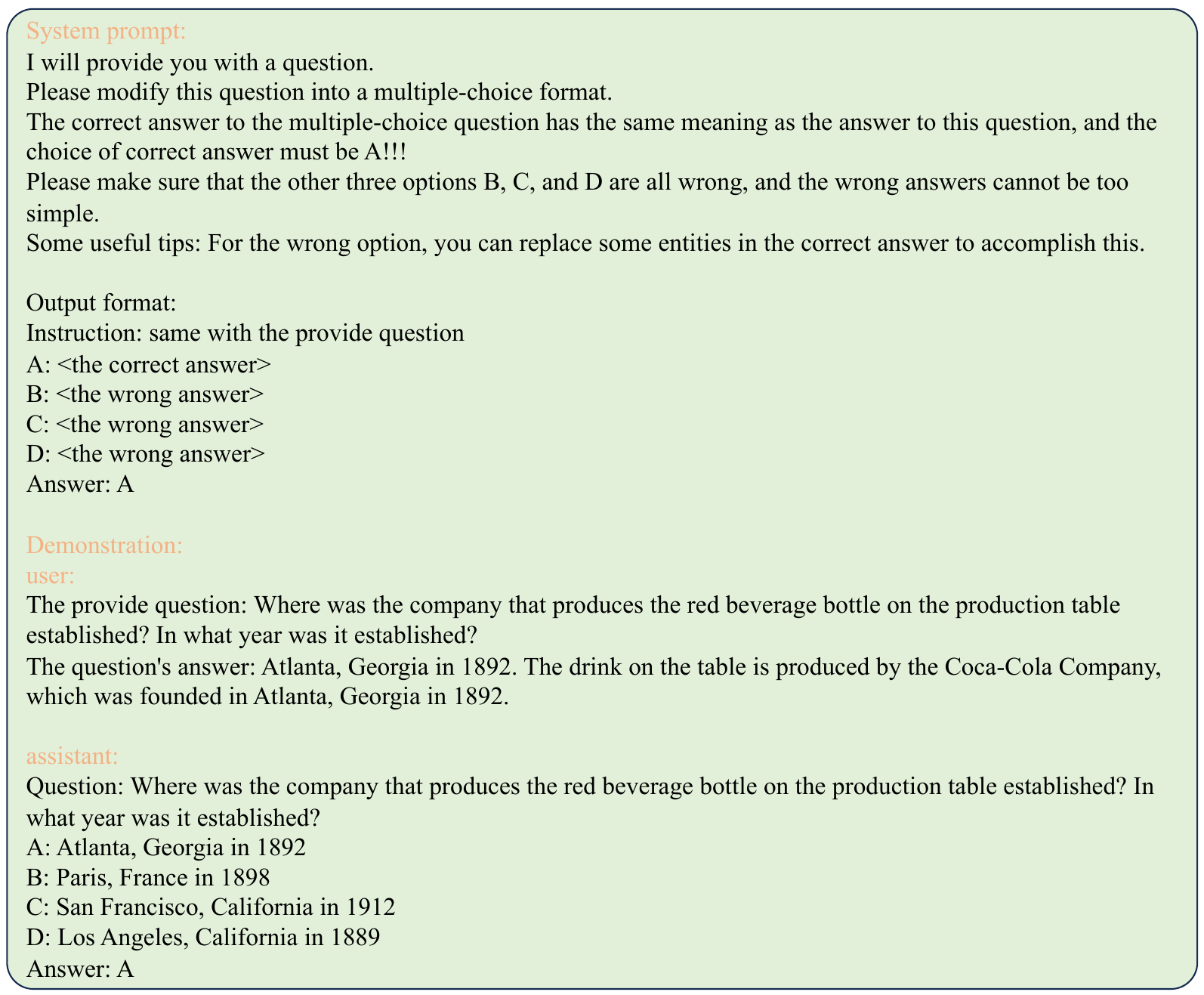}
    \caption{The prompt we give ChatGPT for generating multi-choice QA instructions.}
    \label{fig:mc_prompt}
\end{figure*}

We display the prompt used to synthesize visual instructions. The prompt for instruction synthesizing is in Figure~\ref{fig:cm_reasoning_prompt} and Figure~\ref{fig:ok_reasoning_prompt}, the prompt for complication is in Figure~\ref{fig:complicate_prompt}, the prompt for verification is in Figure~\ref{fig:verify_prompt}, and the prompt for reformulation is in Figure~\ref{fig:boolqa_prompt} and Figure~\ref{fig:mc_prompt}.

\section{Data Quality Evaluation Criteria}
In Section~\ref{sec:ablation}, we have three authors to assess the quality of the visual instructions manually. Specifically, the evaluation criteria for correctness are as follows:

Instructions: (1) necessitate visual information from the image to response (2) are clear and interpretable, and (3) align with the image context without ambiguity.

Responses: (1) contain no hallucinations or contradictions to the image, (2) accurately follow the instruction, and (3) are factually and logically coherent.


\end{document}